\documentclass[conference, 9pt]{IEEEtran}
\IEEEoverridecommandlockouts
\usepackage{lipsum}
\usepackage{cite}
\usepackage{amsmath,amssymb,amsfonts}
\usepackage{algorithmic}
\usepackage{graphicx}
\usepackage{textcomp}
\usepackage{xcolor}
\usepackage{subcaption}
\usepackage{tabularx}
\usepackage{booktabs}
\usepackage{multirow}
\usepackage{graphicx,subcaption,ragged2e}
\usepackage{soul}
\usepackage{tablefootnote}

\begin{document}

\title{AER-LLM: Ambiguity-aware Emotion Recognition Leveraging Large Language Models
}

\makeatletter
\newcommand{\linebreakand}{%
  \end{@IEEEauthorhalign}
  \hfill\mbox{}\par
  \mbox{}\hfill\begin{@IEEEauthorhalign}
}
\makeatother

\author{\IEEEauthorblockN{Xin Hong}
\IEEEauthorblockA{\textit{School of Computing and Information Systems} \\
\textit{University of Melbourne}\\
Melbourne, Australia \\
honxh1@student.unimelb.edu.au}
\and
\IEEEauthorblockN{Yuan Gong\IEEEauthorrefmark{2}}
\IEEEauthorblockA{\textit{CSAIL SLS} \\
\textit{Massachusetts Institute of Technology}\\
Boston, US \\
yuangong@mit.edu} \thanks{\IEEEauthorrefmark{2}Yuan Gong completed this work at MIT and is now with xAI Corp.}
\linebreakand
\IEEEauthorblockN{Vidhyasaharan Sethu}
\IEEEauthorblockA{\textit{School of Electrical Engineering and Telecommunications} \\
\textit{University of New South Wales}\\
Sydney, Australia \\
v.sethu@unsw.edu.au}
\and
\IEEEauthorblockN{Ting Dang}
\IEEEauthorblockA{\textit{School of Computing and Information Systems}\\
\textit{University of Melbourne} \\
%\textit{Cambridge, UK}\\
Melbourne, Australia \\
ting.dang@unimelb.edu.au}
}

% \author{
%     \IEEEauthorblockN{Hong Xin$^1$, Yuan Gong$^2$, Vidhyasaharan Sethu$^3$, Ting Dang$^1$}
%     \IEEEauthorblockA{$^1$ School of Computing and Information Systems, University of Melbourne, Australia}
%     \IEEEauthorblockA{$^2$CSAIL, Massachusetts Institute of Technology, Boston, US}
%     \IEEEauthorblockA{$^3$School of Electrical Engineering and Telecommunications, University of New South Wales, Sydney, Australia
%     \\honxh1@student.unimelb.edu.au, yuan.gong@mit.edu, v.sethu@unsw.edu.au, ting.dang@unimelb.edu.au}
% }

% \author{\IEEEauthorblockN{Hong Xin}
% \IEEEauthorblockA{\textit{School of  Computing and Information Systems} \\
% \textit{University of Melbourne}\\
% Melbourne, Australia \\
% honxh1@student.unimelb.edu.au}
% \and
% \IEEEauthorblockN{Yuan Gong^{*}}
% \IEEEauthorblockA{\textit{CSAIL SLS} \\
% \textit{Massachusetts Institute of Technology}\\
% Boston, US \\
% yuangong@mit.edu} \thanks{$^{*}$Yuan Gong completed this work at MIT and is now with xAI Corp.}
% \linebreakand
% \IEEEauthorblockN{Vidhyasaharan Sethu}
% \IEEEauthorblockA{\textit{School of Electrical Engineering and Telecommunications} \\
% \textit{University of New South Wales}\\
% Sydney, Australia \\
% v.sethu@unsw.edu.au}
% \and
% \IEEEauthorblockN{Ting Dang}
% \IEEEauthorblockA{\textit{School of Computing and Information Systems}\\
% \textit{University of Melbourne} \\
% %\textit{Cambridge, UK}\\
% Melbourne, Australia \\
% ting.dang@unimelb.edu.au}
% }

\maketitle
\addtolength{\topskip}{-0.1cm}  % 调整正文上方的距离

\begin{abstract}
Recent advancements in Large Language Models (LLMs) have demonstrated great success in many Natural Language Processing (NLP) tasks. In addition to their cognitive intelligence, exploring their capabilities in emotional intelligence is also crucial, as it enables more natural and empathetic conversational AI. Recent studies have shown LLMs' capability in recognizing emotions, but they often focus on single emotion labels and overlook the complex and ambiguous nature of human emotions.
This study is the first to address this gap by exploring the potential of LLMs in recognizing ambiguous emotions, leveraging their strong generalization capabilities and in-context learning. We design zero-shot and few-shot prompting and incorporate past dialogue as context information for ambiguous emotion recognition. Experiments conducted using three datasets indicate significant potential for LLMs in recognizing ambiguous emotions, and highlight the substantial benefits of including context information.
Furthermore, our findings indicate that LLMs demonstrate a high degree of effectiveness in recognizing less ambiguous emotions and exhibit potential for identifying more ambiguous emotions, paralleling human perceptual capabilities.

% LLMs struggle more with recognizing more ambiguous emotions and are more effective in recognizing less ambiguous emotions.

% By comparing the performance of LLMs in recognizing ambiguous emotions against human-annotated perceived emotions, we aim to deepen our understanding of how these systems handle emotional subtleties. The ultimate goal is to pave the way for more natural and empathetic communication between humans and computers, thereby bridging the gap between artificial and human intelligence in emotional contexts. The findings of this study have the potential to significantly advance the field of emotion recognition, contributing to the development of more sophisticated and emotionally aware AI systems.

\end{abstract}

{
  \renewcommand{\thefootnote}{}
  % \footnotetext{This work has been submitted to the IEEE for possible publication. Copyright may be transferred without notice, after which this version may no longer be accessible.}
}

\section{Introduction}
Recent advancements in large language models (LLMs)~\cite{santoso2024large} have shown remarkable abilities in comprehending, interpreting, and generating human-like text. This cognitive intelligence facilitates effective human-AI interactions of conversational AI. Equally important is the emotional and social intelligence, which enables it to understand human emotions and adapt its communication accordingly.

% Recognizing and understanding emotion in conversational AI plays a crucial role in facilitating natural communication and offers the potential for continuous monitoring of emotional health conditions through daily interactions. While deep learning has made significant strides in recognizing emotions from various modalities, recent advancements in large language models (LLMs)~\cite{santoso2024large} have demonstrated impressive capabilities to comprehend, interpret, and produce human-like text, substantially pushed the boundaries of emotional intelligence of conversational AI~\cite{lei2023instructerc}. 

Text-based emotion recognition has shown considerable potential, with various feature engineering techniques and advanced deep learning models~\cite{huang2019ana, agrawal2019nelec, bharti2022text}. With the emerging capabilities of LLMs, particularly their proficiency in in-context learning and robust generalization without extensive training, research on exploring their potential for emotion recognition and further use as annotation tools have attracted increasing attentions~\cite{feng2024foundation,niu2024text, tak2023gpt, lei2023instructerc}.  %across various domains without the need for extensive training, relying instead on careful prompt design, such as health monitoring~\cite{}, text report summarisation~\cite{}. 
% Recent studies have also shifted focus towards leveraging LLMs for emotion understanding and demonstrated potential by incorporating conversation histories~\cite{lei2023instructerc}. 
Nonetheless, these investigations predominantly focus on recognizing single emotions, thereby overlooking the complex nature of human emotions. Typically, a single emotion label is obtained from the majority vote of multiple annotators labeling the same stimuli. This approach disregards discrepancies among annotators, which indicates the inherent ambiguity of emotions. %For instance, a sentence perceived as both angry and frustrated reflects a more complex emotional state than one uniformly labeled as angry. These discrepancies, referred to as emotional ambiguity,
High ambiguity, i.e., high disagreement, indicates more complex emotions, and such ambiguity impacts conversations, leading to modified communication strategies and affecting relationship dynamics~\cite{knobloch2010relational}. For example, the listener might use more cautious language and tones or avoid sensitive topics to prevent misunderstandings if high emotional ambiguity is perceived. Future LLMs need to understand the complexity of emotions, recognize emotional ambiguity, and adjust their responses dynamically.

LLMs, trained on diverse and large-scale datasets, enable semantic richness and offer significant potential in comprehending the complexity of emotions. Additionally, their long-range contextual understanding allows them to decode emotions through in-context learning by analyzing conversational history, which is particularly noteworthy. This study aims to explore the potential of LLMs in recognizing inherently ambiguous emotions and the contributions are summarized below:  %Therefore, it is crucial to explore LLMs’ capabilities in understanding emotional ambiguity, especially given their unique in-context learning abilities. 
% This investigation could advance the development of emotional intelligence in LLMs, leading to dynamic conversational AI systems with robust emotional comprehension.

% Emotions are inherently . For example, certain emotions are inherently difficult to recognize, such as frustration, when compared to more straightforward emotions like happiness, thereby exhibiting intrinsic ambiguity. Moreover, certain complex emotions may simultaneously exhibit both surprise and anger. Such ambiguity can significantly impact the conversation and subsequently alter the listener's responses. 

\begin{itemize}
\item This is the first study to analyze LLMs in recognizing ambiguity-aware emotions, demonstrating their potential for human-like emotional intelligence.
\item We proposed zero-shot and few-shot prompt designs and incorporated in-context learning capabilities to enhance recognition, demonstrating an average 35\% relative improvements in terms of Bhattacharyya coefficient.
\item We further included speech features as textual prompt to enhance learning, further enhancing the ambiguous emotion recognition. 
\item Further analysis concerning different levels of ambiguity revealed that LLMs are more effective at recognizing less ambiguous emotions and less effective with more ambiguous ones.
\end{itemize}

% This study aims to explore the potential of LLMs in recognizing inherently ambiguous emotions. By designing prompts using zero-shot and few-shot strategies, and incorporating speech information in textual format alongside conversational text, we seek to enhance LLMs' ability to recognize emotions as discrete distributions that reflect the ambiguity and complexity of human emotions. Our objective is to deepen our understanding of how these systems can identify associated ambiguities and compare their performance with human-annotated perceived emotions. 

This investigation offers valuable insights into emotional intelligence in LLMs and could potentially advance the development of more natural and empathetic conversational AI systems through dynamic emotional responses.

\section{Related work}
A range of feature sets and modeling frameworks have been developed to advance text-based emotion recognition. TF-IDF, which highlights frequent keywords to indicate emotional cues, has been effectively used in emotion classification~\cite{cahyani2021performance}. Subsequently, word embeddings have demonstrated greater effectiveness in capturing semantic relationships, such as Word2Vec, GloVe, and BERT~\cite{huang2019ana, agrawal2019nelec, bharti2022text}. Deep learning models have also advanced, with Bi-Gated Recurrent Units (GRUs), Convolutional Neural Networks (CNNs)~\cite{bharti2022text}, and hierarchical Long Short-Term Memory (LSTM) networks~\cite{huang2019ana} all showing effectiveness. However, their effectiveness is highly dependent on a large annotated dataset for training~\cite{bharti2022text, alswaidan2020survey}.

Recent advances in LLMs, such as LLAMA3~\cite{touvron2023llama} and GPT-4~\cite{achiam2023gpt}, have opened up new possibilities for text understanding and analysis. Their strong generalization capabilities enable effective text comprehension to recognize emotions without the need for extensive retraining~\cite{niu2024text, feng2024foundation, lei2023instructerc, fu2024ckerc, hu2024exploring}. One recent study~\cite{niu2024text} compared the emotion labels recognized by GPT-4 to human annotators and found that labels generated by GPT-4 were preferred by human evaluators. %Additionally, LLMs extended with speech data have demonstrated strong capabilities in inferring emotions solely from speech~\cite{feng2024foundation}. 
Another work proposed InstructERC~\cite{lei2023instructerc}, which treats emotion recognition in conversations as a retrieval-based Seq2Seq paradigm utilizing LLMs to infer emotions from conversation history.
%and present an instruction template suitable for various dialog scenarios. Additionally, it proposes multitasking by recognizing both speaker identity and emotions, thereby implicitly modeling dialogue role relationships to enhance emotional recognition.
 %The state-of-the-art InstructERC proposed by \cite{lei2023instructerc} significantly outperforms previous discriminant models by first identifying speakers and then conducting emotion prediction task using LLMs. 
%Building on this, \cite{fu2024ckerc} further integrated a speaker commonsense identification task to better understand speaker characteristics, further enhancing performance in emotion recognition.

% the enhanced the InstructERC model by exploring implicit information about the speaker. The proposed model, called CKERC, is composed of a speaker commonsense identification task to better understand speaker characteristics and ERC tasks, resulting in competitive performance on popular datasets.

Despite the promises, existing studies all focus on the single emotion label obtained from majority vote and have not studied the inherent ambiguity of emotions. Previous studies on non-LLM models accounting for ambiguity in emotions either use the ambiguity in the loss function as an enhancement for the majority vote recognition~\cite{zhou2022multi} or treat ambiguous emotions as an out-of-distribution (OOD) separate class~\cite{wu2024handling}. The recent study based on LLMs~\cite{niu2024text} includes data with multi-label classes, but the focus is still on single label recognition, and majority of the data is also single-labeled. None of these studies carefully represent emotions with ambiguity or explore the potential of LLMs in recognizing ambiguity-aware emotions, especially their strong capabilities in in-context learning.

\section{Ambiguous emotion recognition via LLMs}

\subsection{System overview}

\begin{figure}
    \centerline{\includegraphics[width=\linewidth]{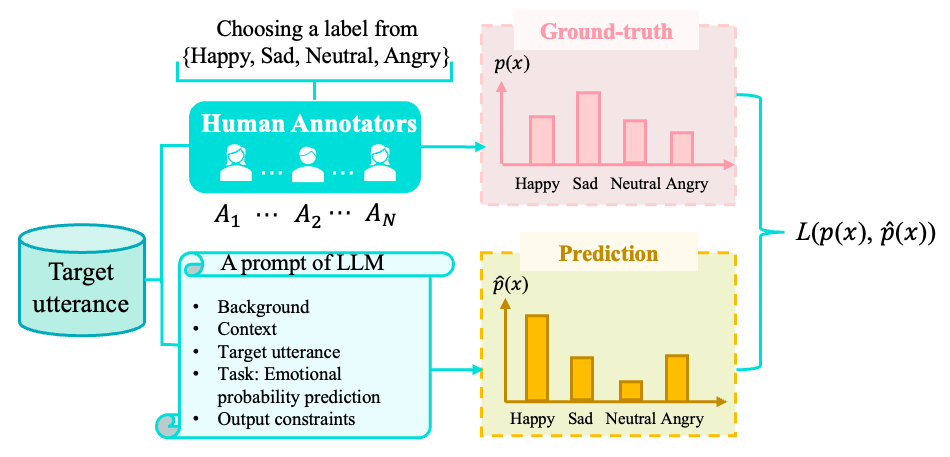}}
    \caption{System overview. $A_i, i \in [1, N]$ represents the $i^{th}$ annotator, $L$ represent the evaluation metrics, including both ambiguity-centric and accuracy-centric metrics. }
    %\td{here ground truth use P(x), and prediction using $\hat{P}(x)$. x is the same but the probability is different. Also you don't need Dp and Dg as you already have p. then in the L, use p(X) instead of Dg, etc.} }
    \label{figure:fm}
    \vspace{-10pt}
\end{figure}

Fig. \ref{figure:fm} illustrates our framework for recognizing ambiguous emotions using LLMs. For each target utterance, we construct detailed prompts for ambiguous emotion recognition.  %that includes background information (BG), context (C) consisting of retrospective dialogue, the target utterance (TU) to be recognized, and the final task of probability prediction with output constraints (OC), guiding the LLMs to generate an emotional probability distribution. Both zero-shot and few-shot prompt have been proposed. 
To evaluate the performance of the LLMs, we compare the predicted emotion distribution $\hat{p}(x)$ with the ground truth distribution $p(x)$, which is inferred from $N$ different human annotators $A_1$ to $A_N$.
%The predicted distribution from the LLM is then compared to the ground-truth distribution using various evaluation metrics, thereby assessing the model's effectiveness in recognizing ambiguous emotions.

\subsection{Prompt design}
\subsubsection{Zero-shot and few-shot prompting}
Carefully designed zero-shot (ZS) and few-shot (FS) prompting are key for LLMs to generalize across various domains~\cite{team2023gemini}. Zero-shot prompting evaluates the capability of pre-trained knowledge in LLMs. Few-shot prompting includes a limited set of demonstration examples within the prompts, facilitating the adaptation of pre-trained knowledge to specific new tasks. %We propose zero-shot and few-shot prompting strategies to learn and predict ambiguous emotional states. 
We design zero-shot prompt as outlined in Table~\ref{tab:prompt} and Eq.~\eqref{eq:prompt}:
\begin{equation}
Prompt_{zs} = BG + C + TU+ Task + OC
    \label{eq:prompt}
\end{equation}

\begin{table}
  \caption{\textcolor{black}{Zero-shot prompt template with context = M}}
  \begin{tabular}{p{2.2cm}|p{5.8cm}}
    \toprule
     & \textbf{Prompt Template}\\
    \midrule
    \textbf{Background (BG)} & Two speakers are talking. \\
    \midrule
    \textbf{Context (M=3)} & The conversation is: \begin{itemize}
        \item Ses01\_F: "We could hide away." 
        \item Ses01\_M: "Run away?"
        \item Ses01\_F: "Mm hmm. We'll build a bunker and never come out." 
        \end{itemize}\\
    \midrule
    \textbf{Target utterance (TU)} & Now Ses01\_M says: "I really don't want to go, I don't want to go..." \\
    \midrule
    
   \textbf{Task} & Predict the probability of the emotion of the sentence from the options [neutral, happy, angry, sad], consider the conversation context.  \\% and audio features.\\
   \midrule
   \textbf{Output constraints (OC)} & Output satisfies the following rules. \begin{itemize}
        \item Rule 1: Generate a dictionary of emotion probabilities in format of \{'neutral':0.1, 'happy':0.0, 'angry':0.1, 'sad':0.8\}. If you think there is only one emotion in the sentence, then give the probability to 1.
        \item Rule 2: Ensure the sum of probability equal to 1.
        \item Rule 3: Do not explain, only need the dictionary.
    \end{itemize} Please check again whether your output follows the three rules.\\
  \bottomrule  
\end{tabular}
\vspace{-5pt}
\label{tab:prompt}
\end{table}

\textbf{Background} (BG) provides information about the conversation scenario, while \textbf{Context} (C) incorporates the retrospective dialogue with $M$ past consecutive utterances. The \textbf{Target Utterance} (TU) indicates the sentence that requires prediction, and the \textbf{Task} presents the question of probability distribution prediction. The final \textbf{Output Constraints} (OC) offers specific instructions to generate the correct form of the distributions. 

In terms of few-shot, we additionally included a few \textbf{Examples (Exps)} in the prompt design, as shown in Table \ref{tab:fs}. The corresponding ambiguous emotion labels in terms of probabilities for these examples are provided to the LLMs for learning. The FS prompting with additional examples is shown in Eq.~\eqref{eq:fs}. %The number of examples matches the context window, and 
\begin{equation}
Prompt_{fs} = BG + C + TU+ \underline{Exps} + Task + OC
    \label{eq:fs}
\end{equation}

\begin{table}
  \caption{\textcolor{black}{Examples in few-shot prompt}}
  \begin{tabular}{p{2.2cm}|p{5.8cm}}
    \toprule
    & \textbf{Prompt Template}\\
    \midrule
   \textbf{Examples (Exps)} %\newline (Same utterance size with context windows) 
   &
    Examples: \begin{itemize}
        \item Sentence 1: Ses01\_F: "We could hide away.". Emotion probabilities: \{'Sadness': 0.33, 'Happiness': 0.33, 'Neutral state': 0.33\}
        \item Sentence 2: Ses01\_M: "Run away?" Emotion probabilities: \{''Sadness': 0.67, 'Neutral state': 0.33\}
        \item Sentence 3: Ses01\_F: "Mm hmm. We’ll build a bunker
and never come out." Emotion probabilities: \{'Sadness': 0.67, 'Happiness': 0.33\}
        \end{itemize} 
        \\
  \bottomrule  
\end{tabular}
\label{tab:fs}
\vspace{-10pt}
\end{table}
% \td{for table 2, find another example here and try to avoid using frustration, but use the four classes we are using now. Also find the examples where the emotion are multicalss most likely, instead of single class. Once you have this example, also make table 1 context consistent with your examples.}
% \tx{done}

\subsubsection{Prompt with speech features}
As humans express emotions through multiple cues, we further included speech in addition to text for ambiguity-emotion recognition. We transformed speech features into text format and incorporated them into the prompt design. Since LLMs have been trained extensively on text data, they are expected to understand speech information in text format, e.g., high pitch values: 4. Specifically, we extracted 88-dimensional eGeMAPS features ~\cite{eyben2015geneva}, a standard acoustic parameter set.
It described the speech features of the target utterance in textual format, as illustrated in Table~\ref{tab:audio}, with the full prompts shown in Eq.~\eqref{eq:audio} and~\eqref{eq:audio1}.
\begin{equation}
\small Prompt_{zs}^a = BG + C + TU+ \underline{Speech}  + Task + OR
    \label{eq:audio}
\end{equation}
\begin{equation}
\small Prompt_{fs}^a = BG + C  + TU+ \underline{Speech} + Task + \underline{Exps} + OR 
    \label{eq:audio1}
\end{equation}

\begin{table}[t]
  \caption{Speech features in textual format in prompt}
  \begin{tabular}{p{1.7cm}|p{6.3cm}}
    \toprule
    & \textbf{Prompt Template}\\
    \midrule
   \textbf{Speech Features} & Here are 88 speech features of the current speaker's sentence.
    The features are: 
   Average Fundamental Frequency in Semitones from 27.5 Hz: 37.039505 ...  \\
    % F0semitoneFrom27.5Hz\_sma3nz\_amean: 37.039505 ... \\
  \bottomrule  
\end{tabular}
\label{tab:audio}
\vspace{-10pt}
\end{table}

\vspace{-2pt}
\subsection{Context-aware recognition}
A major advantage of LLMs lies in their capability for in-context learning, offering the potential to analyze long past conversation histories for emotion recognition. Since emotions generally evolve smoothly within dynamic conversations, considering past conversations provides a more comprehensive understanding of the emotional state over time. A longer context window enables LLMs to effectively decode information over extended ranges, and we formally study this by increasing the number $M$ of context windows in the prompt design, i.e., including the corresponding text information from the past $M$ utterances, %with $M$ ranging from 0 to 30, 
and evaluated the performance accordingly.

\section{Experimental setup and results}
\subsection{Experimental setup}
\subsubsection{Dataset} 
Three datasets are used, MSP-Podcast~\cite{martinez2020msp}, IEMOCAP~\cite{busso2008iemocap} and GoEmotions~\cite{demszky2020goemotions}.
MSP-Podcast contains recordings cover a wide range of subjects in Podcasts. We selected four emotional labels: neutral, angry, happy, and sad. Any utterances annotated by annotators outside of these four categories will be excluded from the analysis, leading to 4114 utterances. We manually reorganized sentences into the original full podcast to enable dynamic information. For IEMOCAP, we also selected the same four emotional labels, resulting in a total of 4370 utterances. Examples in few-shot prompting are from the same session with the target utterance.

% Each utterance was evaluated by three annotators. %, who can select more than one emotion from a predefined set. 
% We selected four emotional labels: neutral, angry, happy, and sad, with 'excited' merged into happy. Any utterances annotated by annotators outside of these four categories will be excluded from the analysis, %Different from labels by majority vote, any individual annotation within the three annotations for a single utterance that lies outside of these four types will not be included, 
% resulting in a total of 4370 utterances.

GoEmotions is a text-based dataset sourced from Reddit. %annotated with 27 distinct emotion categories plus a Neutral label. 
Given the long-tail distribution of emotion labels,  %might not effectively demonstrate the validity of our model, 
we selected the 4 most common labels (admiration, gratitude, approval, and amusement) along with Neutral. To ensure adequate representation of ambiguous emotions (more than one label per utterance), we applied log inverse frequency weighting and selected 210 instances, with 33\% being multi-labeled.\footnote{GoEmotions contains only single utterances without context information. We randomly sampled specific sentences as examples for few-shot prompting.}
% \tx{We randomly sampled specific sentences as examples for all utterances.}

\subsubsection{Models} Gemini-1.5-Flash was chosen as the LLM backbone due to its capability for processing long-range contexts, with a context window of up to one million tokens. The experiments were conducted using the Gemini API~\cite{reid2024gemini}. For few-shot prompts, we tested 5 and 10 examples in GoEmotions. In MSP-Podcast and IEMOCAP, we matched the number of few-shot examples to the context window, varying the context window within [0,30]\footnote{Code link: https://github.com/mHealthUnimelb/AER-LLM.}. %simulating practical scenarios where emotions of the past $M$ utterances are accurately recognized and used to infer subsequent emotions. We varied the context window $M$ from 0 to 30.

\subsubsection{Evaluations}
We include both uncertainty-centric and accuracy-centric metrics to evaluate our model's performance. For uncertainty-centric metrics, we use Jensen-Shannon Divergence (JS), Bhattacharyya coefficient (BC), and $R^2$. JS divergence measures the difference between predicted and ground truth distributions, with lower values indicating better predictions. BC measures the similarity between these distributions, and $R^2$ assesses the goodness-of-fit, with higher values indicating better performance. Additionally, we estimate Expected Calibration Error (ECE), where smaller values denote better calibration of probabilistic predictions. All four metrics range within [0,1]. For accuracy-based metrics, we obtain a single label from the predicted distribution by selecting the maximum probability and compared to the majority vote of the labels to estimate accuracy, F1-score and unweighted average recall (UAR). %This helps assess the model's performance on majority vote prediction, even if the prompt is not specifically designed for it.

\subsubsection{Baseline descriptions}
%We selected these baselines closely match our experimental settings and ensure a fair comparison. 
Baseline~\cite{aldeneh2021you} utilizes pretrained embeddings for single emotion recognition, while~\cite{wu2023distribution} treats ambiguous emotion sentences as additional out-of-distribution class for ambiguous emotion recognition. The other three studies~\cite{feng2024foundation, niu2024text, lei2023instructerc} focus on LLMs for single emotion recognition tasks, with~\cite{lei2023instructerc} building a two-step system.% and~\cite{niu2024text} comparing results with human evaluators.

%  Pretrained~\cite{aldeneh2021you} utilized speaker embeddings representing the speech characteristics for emotion detections. 
% Multi-LLMs~\cite{feng2024foundation} ensembled the outputs from multiple LLMs to mitigate the biases in language reasoning.
% InstructERC~\cite{lei2023instructerc} explored zero-shot prompting with LLMs for the speaker's identification task and emotions recolonization task.
% Pretrained (W2V2 + Bert)~\cite{wu2023distribution} focused on modelling emotion ambiguity by minimising the loss of KL and DPN from emotion distributions. 
% GPT-4 \cite{niu2024text} compared the emotion labels recognized by GPT-4 to human annotators. % and found that labels generated by GPT-4 were preferred by human evaluators.

\vspace{-2pt}
\subsection{Performance on ambiguity-aware prediction}
% \td{maybe here we just include text-based results for zero-shot and few-shot? then next section compares text and text + audio? then context comparison; }
% \textbf{Baselines:} Bert. InstructECR
% \begin{table}
% \centering
%   \caption{\textcolor{black}{Performance of ambiguity-aware emotion prediction with context window = 20}\td{discuss the baselines carefully}}
%   \begin{tabular}{llllll}
%     \toprule
%     Method & KL\textsubscript{std} & BC & R$^2$ & ECE \\
%     \midrule
%     % Bert~\cite{wu2023distribution}& - & - & - & - & 78.16 & - \\
%     % InstructECR~\cite{lei2023instructerc}\footnote{The data is slightly different due to the different tasks. } & - &- &- &- & - & 53.38 \\
    
%     Zero-shot (only text) & 49.59 & 0.46 & 0.49 & 56.2 \\
%     Zero-shot (text+audio)& 43.76 & 0.51 &  0.51 & 51.21  \\
%     Few-shot (only text)  & 30.49 & 0.67 & 0.58 & 30.73 \\
%     Few-shot (text+audio)  & 27.76 & 0.69 & 0.59 & 28.88\\
%   \bottomrule  
% \end{tabular}
% \label{tab:performance}
% \end{table}

\begin{table}
\centering
  \caption{Performance on ambiguity-aware emotion prediction. T represents text and S represents speech.} %with context window = 20}
  \begin{tabular}{l|cc|ccc|l}
    \toprule
    & Modality & Prompt & JS\(\downarrow\) & BC\(\uparrow\) & R\(^2\)\(\uparrow\) & ECE\(\downarrow\) \\
    \midrule
    \cmidrule{1-7}

     \multirow{4}{*}{MSP-Podcast} & \multirow{2}{*}{T} & ZS & 0.56 & 0.39 & 0.41 & 0.62 \\
%        && Zero-context  & 12.75 & 0.39 & 0.40 & 63.01 \\
    && FS& 0.42 &0.58& 0.54& 0.42 \\
    \cmidrule{2-7}
    &\multirow{2}{*}{T+S} & ZS & 0.45 &0.54&0.52& 0.47  \\
    && FS  & \textbf{0.40}& \textbf{0.61}& \textbf{0.56}& \textbf{0.40}\\
    \midrule
    \multirow{4}{*}{IEMOCAP} & \multirow{2}{*}{T} & ZS & 0.51 & 0.46 & 0.49 & 0.56 \\
%        && Zero-context  & 12.75 & 0.39 & 0.40 & 63.01 \\
    && FS& 0.37 & 0.67 & 0.58 & 0.30 \\
    \cmidrule{2-7}
    &\multirow{2}{*}{T+S} & ZS & 0.47 & 0.51 &  0.51 & 0.51  \\
    && FS  & \textbf{0.35} & \textbf{0.69} & \textbf{0.59} & \textbf{0.29}\\
    \midrule
    \multirow{2}{*}{GoEmotions} &\multirow{2}{*}{T}& ZS&  0.49 & 0.54 & 0.43 & 0.47 \\
     & & FS&  \textbf{0.44} & \textbf{0.60} & \textbf{0.48} & \textbf{0.39} \\ %5-shot
  \bottomrule  
\end{tabular}
\vspace{-10pt}
\label{tab:performance}
\end{table}

% \begin{table}
% \centering
%   \caption{Performance on ambiguity-aware emotion prediction. T represents text and S represents speech.} %with context window = 20}
%   \begin{tabular}{l|cc|ccc|l}
%     \toprule
%     & Modality & Prompt & JS\downarrow & BC\uparrow & R\[^2\]\uparrow & ECE\downarrow \\
%     \midrule
%     \multirow{4}{*}{MSP-Podcast} & \multirow{2}{*}{T} & ZS & 0.56 & 0.39 & 0.41 & 0.62 \\
% %        && Zero-context  & 12.75 & 0.39 & 0.40 & 63.01 \\
%     && FS& 0.42 &0.58& 0.54& 0.42 \\
%     \cmidrule{2-7}
%     &\multirow{2}{*}{T+S} & ZS & 0.45 &0.54&0.52& 0.47  \\
%     && FS  & \textbf{0.40}& \textbf{0.61}& \textbf{0.56}& \textbf{0.40}\\
%     \midrule
%     \multirow{4}{*}{IEMOCAP} & \multirow{2}{*}{T} & ZS & 0.51 & 0.46 & 0.49 & 0.56 \\
% %        && Zero-context  & 12.75 & 0.39 & 0.40 & 63.01 \\

%     && FS& 0.37 & 0.67 & 0.58 & 0.30 \\
%     \cmidrule{2-7}
%     &\multirow{2}{*}{T+S} & ZS & 0.47 & 0.51 &  0.51 & 0.51  \\

%     % & Few-shot (only text)  & 30.49 & 0.67 & 0.58 & 30.73 \\
%     && FS  & \textbf{0.35} & \textbf{0.69} & \textbf{0.59} & \textbf{0.29}\\
%     \midrule
%     \multirow{2}{*}{GoEmotions} &\multirow{2}{*}{T}& ZS&  0.49 & 0.54 & 0.43 & 0.47 \\
%      & & FS&  \textbf{0.44} & \textbf{0.60} & \textbf{0.48} & \textbf{0.39} \\ %5-shot
%     % Goemotion & Text & 10-shot&  8.99 & 0.58 & 0.48 & 39.05 \\
%      % Goemotion & Text & 15-shot&  8.11 & 0.60 & 0.49 & 36.95 \\
%   \bottomrule  
% \end{tabular}
% \vspace{-10pt}
% \label{tab:performance}
% \end{table}

Table~\ref{tab:performance} demonstrates the performance on three datasets. For the text modality, the zero-shot prompt achieves relatively acceptable performance, e.g., $R^2$ of 0.41 for MSP-Podcast, 0.49 for IEMOCAP, and 0.43 for GoEmotions. With few-shot prompting, it demonstrates significant improvement, with approximately 25\% reduction in JS, 49\% increase in BC, and 31\% increase in \(R^2\) for MSP-Podcast. Similar increasing trend is also observed in IEMOCAP and GoEmotions. This suggests that LLMs are strong in learning from few examples for ambiguous emotion recognition, and leveraging their in-context learning capabilities by looking at past examples is highly beneficial.

For the joint modeling of text and speech modality, both zero-shot and few-shot demonstrate significant improvement compared to the corresponding performance using text only, suggesting the LLMs' capabilities in recognizing speech information despite being in textual format. Furthermore, the few-shot prompting consistently outperforms the zero-shot prompting, exhibiting superior performance for all three datasets. %with 0.40, 0.58, and 0.54 for JS, BC and \(R^2\) respectively in MSP-Podcast. 
The consistent trend in ECE also implies a similar effect on probability calibrations, suggesting an improved interpretation of ambiguity. The disparities are stable because LLM typically provide consistent responses. 

% The standard deviations of LLM in the best setting across five runs for JS, BC, $R^2$ and ECE are 1e-6, 2e-6, 1e-6, and 4e-6 respectively. or put this into TABLE IV's subscript?

% We further compared the performance between improvised and scripted partitions in IEMOCAP using the optimal settings. While the improvised partitions utilize free text, offering greater natural richness, the scripted partitions adhere to predefined scripts and therefore represent less natural settings. As shown Table~\ref{tab:spim}, the improvised settings show significantly better performance, possibly due to the more natural emotions expressed and the fact that LLMs are trained on more natural text.

% \begin{table}
% \centering
%   \caption{\textcolor{black}{Performance comparison on scripted and improvised data in IEMOCAP.} }
%   \begin{tabular}{l|lll|l}
%     \toprule
%     Dataset\footnote{Utterances without a majority vote label were removed for the comparison.} & JS\downarrow & BC\uparrow & R$^2$\uparrow  & ECE\downarrow \\ %& ACC\uparrow  & W-F1\uparrow  \\ 
%     \midrule
%     Script & 0.46 & 0.56 & 0.46 & 39.99\\ %& 41.25 & 38.75 \\ 
%     % \midrule
%     Impro & 0.26 & 0.80 & 0.69 & 15.46\\ %& 71.76 & 72.34 \\
%   \bottomrule  
% \end{tabular}
% \label{tab:spim}
% \end{table}

\vspace{-2pt}
\subsection{Impact of context window}

Fig. \ref{fig:winds} shows the few-shot performance using text and speech with the increasing context windows from 0 to 30 in MSP-Podcast. Including context information proves significantly beneficial compared to that without contextual information. %Additionally, increasing the context window size generally enhances model performance. 
When the window size increases from 0 to 5, we observe a 16\%, 28\%, 22\% and 19\% improvements in terms of JS, BC, $R^2$ and ECE, respectively. As the context window expands beyond 10 to 30, the observed improvements become marginal, indicating that further increasing the context window size beyond 10 may not substantially enhance ambiguous emotion recognition. In the IEMOCAP dataset, the most benefit was obtained when the context window increases to 20. These findings collectively suggest that incorporating context information is crucial for ambiguous emotion recognition in LLMs, and a context window of 10 to 20 is likely to be adequate. This is reasonable, as humans do not need infinite context to recognize emotion.
%past 20 historical utterances could be adequate for LLMs to understand the context to recognize the current ambiguous emotion. % reach their peaks, while the BC values slightly increase until the window size reaches 30.

% Fig. \ref{fig:winds} shows the performance with the increasing of context windows from 5 to 30. It is evident that increasing the context window size generally enhances model performance across all metrics. From window size increases from 5 to 10, KL\textsubscript{std} and accuracy improve by 3\% and BC increases by around 5\%. After that, as historical windows increasing to 20, both KL\textsubscript{std} and accuracy reaches their peaks, while the BC values slightly increase until window size reaches 30. 

\begin{figure}    \centering{\includegraphics[width=0.9\linewidth]{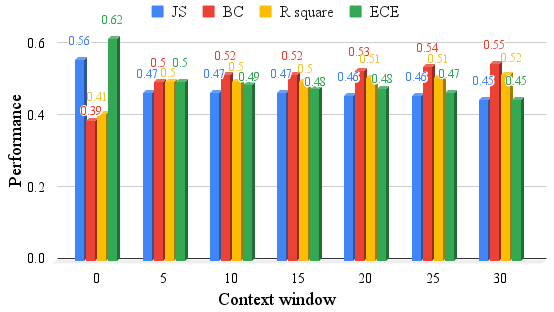}}
    \caption{Performance comparison with increasing context windows using text and speech in MSP-Podcast.}  %on adjusting Historical windows, keeping other settings the same (text+aduio, few-shot prompting)}
    \label{fig:winds}
    \vspace{-10pt}
\end{figure}

\begin{figure}[t]
\centerline{\includegraphics[width=\linewidth]{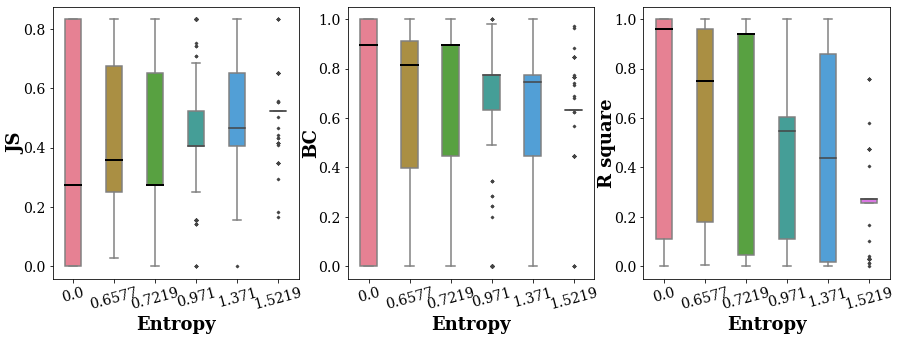}}
    \caption{Performance comparison among different levels of ambiguity in MSP-Podcast. A small entropy indicates less ambiguous emotion.}
    \vspace{-12pt}
    %Box plots comparing entropy and ambiguity evaluations; Experiments were conducted with a context window of 20, using both audio and text features in few-shot prompting; When the entropy is 0, 0.9149, and 1.5835, there are 2,410, 1,813, and 147 instances, respectively.}
    \label{fig:boxplot}
\end{figure}

\vspace{-2pt}
\subsection{Performance with different ambiguity levels}
To gain deeper insights into how LLMs recognize varying levels of ambiguity in emotions, we evaluated their performance with respect to different ambiguity levels. We used entropy, inferred from the ground truth distributions, as an indicator of the ambiguity levels. As entropy increases, the utterance exhibits greater ambiguity in emotion. An entropy of 0 indicates unanimous agreement on one emotion class. As shown in Fig.~\ref{fig:boxplot}, six majority entropy with each more than 100 utterances are shown. %An entropy of 0.6577, 0.7219, 0.9710 indicates annotators assigned emotions into two classes, while an entropy of 1.5219 signifies that three labels were provided.

%IEMOCAP:  As entropy increases, the utterance exhibits greater ambiguity in emotion. As shown in Fig.~\ref{fig:boxplot}, an entropy of 0 in the x axis indicates unanimous agreement on one emotion class. An entropy of 0.9149 shows two out of three annotators agreed, while 1.5835 means all three assigned different labels. 

% Fig. \ref{fig:boxplot} depicts a set of box plots with three evaluation metrics spreading on three groups of entropy. As entropy increases, the utterance contains more ambiguous emotions: An entropy of 0 indicates unanimous agreement among annotators on a single class; an entropy of 0.9149 reflects that two annotators agreed on a class while the third chose a different emotion; an entropy of 1.5835 suggests that all three annotators provided different labels. We have findings as following: 
As entropy increases, the medians (black lines) of JS rise, while BC and R$^2$ decrease in general, except when entropy is 0.7219. The trend indicates that LLMs are more effective at recognizing less ambiguous emotions. This is likely due to the inherent complexity of predicting a high entropy emotional distribution. This observation aligns with the human difficulty in recognizing more ambiguous emotions~\cite{lee2020perceiving}. Better performance is observed with an entropy of 0.7219, as we found that 92\% of the utterances are annotated with at least a neutral class, more than in neighbouring entropy groups, and LLMs recognize neutral with a 75\% true positive rate, leading to higher performance.

\vspace{-2pt}
\subsection{Prediction on majority vote}
We further estimated the single emotion from the distribution by selecting the one with the highest probability and compared it with the majority vote. Noted that the prompt design for LLMs is not optimized for recognizing the majority vote, this analysis is designed to provide insights into LLMs capabilities in understanding the most likely emotion. It is not a direct comparison due to the slightly different data and tasks.
% \tx{FYI: Previously not sure whether to change baselines. Reasons of remaining these two: W2V2+Bert is the latest paper (2022) with better performance than papers on 2020/2019; InstructERC (using LLMs), submitted in 2023, has been cited 49 times.} 

In Table~\ref{tab:acc}, the best performances on MSP-Podcast, IEMOCAP and GoEmotions achieve 56.15\%, 59.06\% and 51.05\% in terms of W-F1, respectively, demonstrating emotional understanding of LLMs. Note that the prompt is not designed for single-label emotion recognition and LLMs is not trained, but it still achieves comparable performance to the models specifically trained for single-label emotion recognition.
The accuracy with respect to the context window and the entropy in MSP-Podcast is further demonstrated in Fig.~\ref{fig:WF1}. Notably, there is no majority label when entropy is 1.5219 as two classes share the same probabilities. Compared to a context window of 0, a context window of 30 achieved higher accuracy across all entropy groups. Additionally, better performance is observed when entropy is 0, indicating no ambiguity, compared to high entropy of 1.371, which corresponds to high ambiguity. %
We also observe a similar pattern in IEMOCAP, with a more consistent increasing trend as the context window increases and a decreasing trend as the entropy increases. %: high contect window  consistent increasing increasing context window can boost performance for LLMs even when the text is more ambiguous.
% IEMOCAP: A context window of 20 achieved the highest accuracy, and better performance is observed in low ambiguity regions compared to high ambiguity ones. Notably, there is no majority label when all annotators assign different labels.\td{discussion needs to be changed as well. After discussing this, we can say we also find the similar pattern in IEMOCAP.}

% As shown in Table \ref{tab:acc}, \td{discussion on new baselines}the novel W2V2 + Bert model \cite{wu2023distribution} achieved a strong performance in previous work, with an accuracy of 78.16, significantly higher than our LLM-based approaches. Additionally, InstructERC \cite{lei2023instructerc}, powered by LLMs achieved a W-F1 score of 53.38, outperforming our models in this metric. 

% Baseline description: 
% W2V2 + Bert  proposed by \cite{wu2023distribution} addressed on modelling emotion ambiguity by minimising the loss of KL and DPN from emotion distributions (DPN-KL system). 

% InstructERC \cite{lei2023instructerc} experimented a series of find-tuned LLMs (ChatGPT3.5 and Llama, Llama2) on adapted LoRA \cite{hu2021lora} or zero-shot prompting, to conduct the speaker's identification task and emotions recolonization task. We selected the ChatGPT3.5 with zero-shot prompting on the InstructERC framework to make a fair comparison with our models. 

\begin{table}
\centering
  \caption{Performance on majority vote prediction, selected as the class with the maximum probability in the predicted distribution.} %\tablefootnote{It is not a direct comparison due to the slightly different data and tasks.}.}%\tx{Instances without majority voting labels are removed.}} 
  \begin{tabular}{l|cc|ccc}
    \toprule
     & Modality & Methods & ACC & W-F1 & UAR \\
    \midrule
        \multirow{6}{*}{MSP-Podcast} 
    &\multirow{2}{*}{T}&ZS  &35.24 &42.78&46.46\\
    &&FS& \textbf{50}&\textbf{53.23}&\textbf{48.31}\\
    \cmidrule{2-6}
     &\multirow{4}{*}{T+S} &  Pretrained~\cite{aldeneh2021you}& -&-&50.0 \\
    & &MLLMs~\cite{feng2024foundation} & - &- &\textbf{52.59}\\
    \cmidrule{3-6}
    &&ZS&50.55 &50.4& 44.21\\
    &&FS&\textbf{55.58}&\textbf{56.15}&46.88\\
    \midrule
    \multirow{7}{*}{IEMOCAP} & 
    
    \multirow{4}{*}{T} &InstructERC~\cite{lei2023instructerc}& - & 53.38 & -\\
    \cmidrule{3-6}
    &&ZS  & 43.36 & 42.92 & 52.25\\
        % &&Zero-context& 33.32 & 34.16 \\

    &&FS& \textbf{57.87} & \textbf{58.43} & \textbf{65.39}\\
    
    \cmidrule{2-6}
        & \multirow{4}{*}{T+S}& MLLMs~\cite{feng2024foundation} & - & - & 50.36 \\
        && Pretrained~\cite{wu2023distribution} & \textbf{78.12} & - & -\\ 
    \cmidrule{3-6}
    &&ZS& 48.12 & 49.18 & 55.20 \\
    &&FS&58.75& \textbf{59.06} & \textbf{65.68}\\
    \midrule
    \multirow{3}{*}{GoEmotion}
    &\multirow{3}{*}{T} & GPT-4 \cite{niu2024text}& - & - & 48.5 \\
    \cmidrule{3-6}
    &  &ZS & 37.14 & 35.68 & 44.74 \\
    % Goemotion&Text &5-shot & 46.19 & 45.61 \\
    % Goemotion&Text &10-shot & 47.62 & 47.65 \\
    & &FS & \textbf{50.48} & \textbf{51.05} & \textbf{52.98}\\ %15
  \bottomrule  
\end{tabular}
\vspace{-5pt}
\label{tab:acc}
\end{table}
% \footnotetext{Comparisons to baselines are indirect due to differing datasets in our ambiguous emotion study.}

% \begin{figure}[t]
%     \centerline{\includegraphics[width=\linewidth]{IV-E-acc.png}}
%     \caption{Accuracy on adjusting context windows across two entropy groups with other settings held constant (text+audio, few-shot prompting). The entropy group with a value of 1.5835 is not included in the bar chart due to the absence of majority voting labels.}
%     \label{fig:acc}
% \end{figure}

\begin{figure}[t]
\centerline{\includegraphics[width=0.85\linewidth]{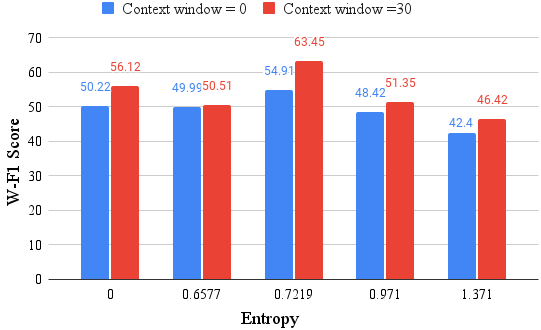}}
    \caption{Comparison of W-F1 across five entropy groups with context window = 0 and 30 using MSP-Podcast.}
    \label{fig:WF1}
    \vspace{-15pt}
\end{figure}

% \begin{figure}[t]
%     \centerline{\includegraphics[width=\linewidth]{impro_boxplot.png}}
%     \caption{[Impro dataset]Box plots comparing entropy and ambiguity evaluations. When the entropy is 0, 0.9149, and 1.5835, there are 1486, 936, and 26 instances, respectively.}
%     \label{fig:imbp}
% \end{figure}

% \begin{figure}[t]
%     \centerline{\includegraphics[width=\linewidth]{script_boxplot.png}}
%     \caption{[Script dataset]Box plots comparing entropy and ambiguity evaluations. When the entropy is 0, 0.9149, and 1.5835, there are 1153, 749, and 20 instances, respectively.}
%     \label{fig:spbp}
% \end{figure}
\vspace{-2pt}
\section{Conclusion}
\vspace{-2pt}
We investigated LLMs potential in recognizing ambiguous emotions and discovered that it can comprehend such emotions to a certain extent without additional training. Incorporating previous dialogues by leveraging LLMs in-context learning capabilities significantly enhances its emotional intelligence, with a context window of 10 to 20 being adequate.
Moreover, these models demonstrate greater proficiency in recognizing less ambiguous emotions compared to highly ambiguous ones, similar to human perception. These findings highlight the potential of LLMs for application in emotional conversational AI.

%Further exploration is needed to extend the analysis to more real-world settings, incorporating a wider range of emotion categories and more realistic daily life scenarios. This will help validate the model's potential and robustness in generalizing across diverse conversational AI contexts.

\bibliographystyle{ieeetr}
\bibliography{refs}

\newpage

\end{document}